\documentclass[runningheads]{llncs}

\usepackage{eccv}
\usepackage{algorithm}
\usepackage{algpseudocode}
\usepackage{multirow}

\usepackage{eccvabbrv}

\usepackage{graphicx}
\usepackage{booktabs}
\usepackage{multirow,multicol}
\usepackage{tabularx}
\usepackage{marvosym}
\usepackage[accsupp]{axessibility}  

\makeatletter
\def\thanks#1{\protected@xdef\@thanks{\@thanks
        \protect\footnotetext{#1}}}
\makeatother
\usepackage{hyperref}

\usepackage{orcidlink}

\begin{document}

\title{SyncCache: Exploiting Asymmetric Dynamics for Fast Audio-Driven Portrait Animation} 

\titlerunning{SyncCache}


\author{
Juncheng Ma\inst{1}\orcidlink{0009-0001-9027-3111} \and
Yuxuan Du\inst{1}\orcidlink{0009-0002-7529-6435} \and
Yanan SUN\inst{2}\orcidlink{0000-0002-1369-2902} \and
Zhening Xing\inst{2}\orcidlink{0000-0002-6092-0891} \and \\
Changlin Li\inst{3}\orcidlink{0000-0002-9565-8790} \and
Zhenyu Tang\inst{1}\orcidlink{0009-0000-8018-1849} \and
Bo Li\inst{4}\orcidlink{0000-0001-7817-0665} \and
Peng-Tao Jiang\inst{4}\orcidlink{0000-0002-1786-4943} \and
Li Yuan\inst{1}\orcidlink{0000-0002-2120-5588} \and \\
Daquan Zhou\textsuperscript{\Letter}\inst{1}\orcidlink{0000-0002-4771-1796} \and
Yonghong Tian\textsuperscript{\Letter}\inst{1}\orcidlink{0000-0002-2978-5935}
\thanks{\textsuperscript{\Letter}Corresponding author.}
}

\authorrunning{J. Ma et al.}

\institute{
Shenzhen Graduate School, Peking University, China, \\
\email{junchengma25@stu.pku.edu.cn} \\
\and
Shanghai AI Laboratory, China, 
\and
Tencent Hunyuan, China,
\and
vivo, China
}

\maketitle

\begin{abstract}
Diffusion Transformers (DiTs) have significantly advanced audio-driven portrait animation, but their high computational cost leads to substantial inference latency. Although training-free diffusion caching accelerates inference significant, existing methods are primarily developed for text-conditioned generation and overlook the spatial and modality imbalances inherent in audio-driven portrait animation. In this paper, we propose \textbf{SyncCache}, a training-free caching acceleration method tailored for DiT-based portrait animation that explicitly exploits asymmetric dynamics. Specifically, high-frequency dynamics driven by audio conditions and concentrated in human regions are more challenging and critical to cache and reuse than the low-frequency visual background in portrait animation. First, we introduce Spatially-Asymmetric Probing to prioritize error sensitivity in dynamic human region. Second, through Modality-Decoupled Caching, we bypass heavy DiT block by reusing stable inter-block residuals, while continuously recomputing lightweight audio blocks to preserve precise lip synchronization. Furthermore, we introduce a cache ratio to control cache capacity and formulate memory-adaptive cache selection as an offline dynamic programming problem without online overhead. Extensive experiments demonstrate that SyncCache achieves superior speed–quality trade-offs, delivering up to $4.12\times$ acceleration on HunyuanVideo-Avatar and $3.75\times$ on Wan-S2V  with near-lossless visual fidelity and precise audio alignment.
\keywords{Diffusion Caching \and Audio-driven Portrait Animation \and Inference Acceleration}
\end{abstract}    
\section{Introduction}
\label{sec:intro}

\begin{figure}[t]
  \centering
  \includegraphics[width=1\linewidth]{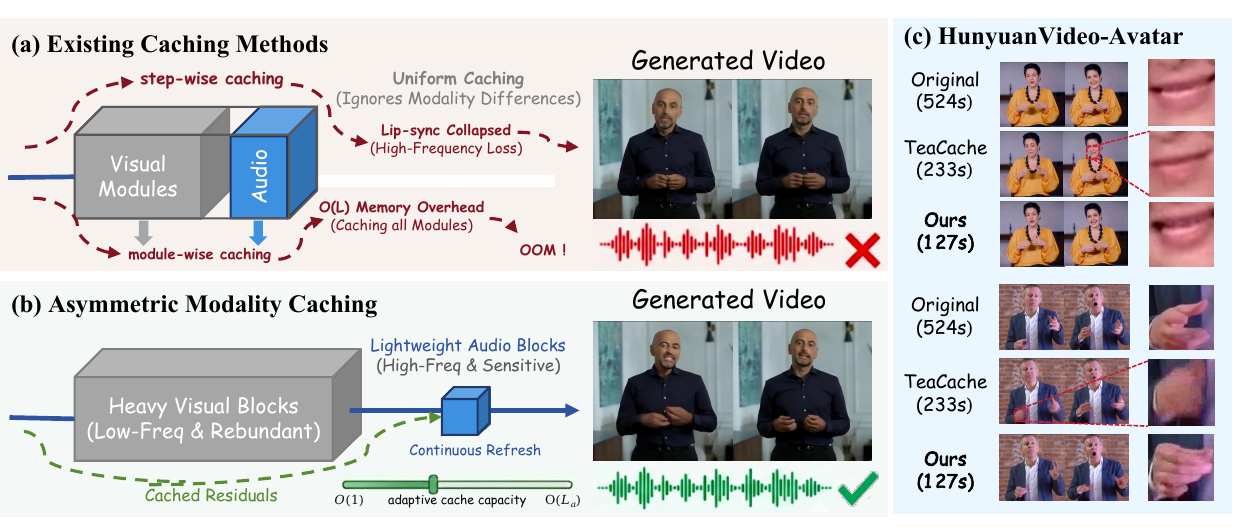}
  \caption{\textbf{Comparison of Caching Paradigms.} \textbf{(a)} Existing methods blindly assume uniform modality dynamics, leading to either catastrophic high-frequency loss or severe memory overhead. \textbf{(b)} SyncCache explicitly exploits asymmetric modality dynamics, bypassing heavy visual computations via stable residuals while continuously refreshing lightweight audio blocks. \textbf{(c)} Consequently, SyncCache achieves massive acceleration (e.g., 4.12$\times$) while  preserving delicate high-frequency details that collapse in baselines.}
  \label{fig:teaser}
\end{figure}
Diffusion Transformers (DiTs) \cite{peebles2023scalable} have advanced video generation, demonstrating exceptional scalability and visual fidelity. Recently, this architecture have been extended to audio-driven portrait animation~\cite{wang2025fantasytalking,gan2025omniavatar,cui2025hallo3}, enabling the synthesis of realistic talking-face videos from an identity reference and an audio clip. Despite remarkable advances, inference latency remains a critical bottleneck. Generating a short clip requires several minutes even on multi-GPU systems (e.g., generating a 15-second video with HunyuanVideo-Avatar \cite{chen2025hunyuanvideo} takes approximately 10 minutes on 8$\times$A800), severely hindering practical creative iteration.

\begin{figure}[tb]
  \centering
  \begin{subfigure}{0.55\linewidth}
    \includegraphics[width=1\linewidth]{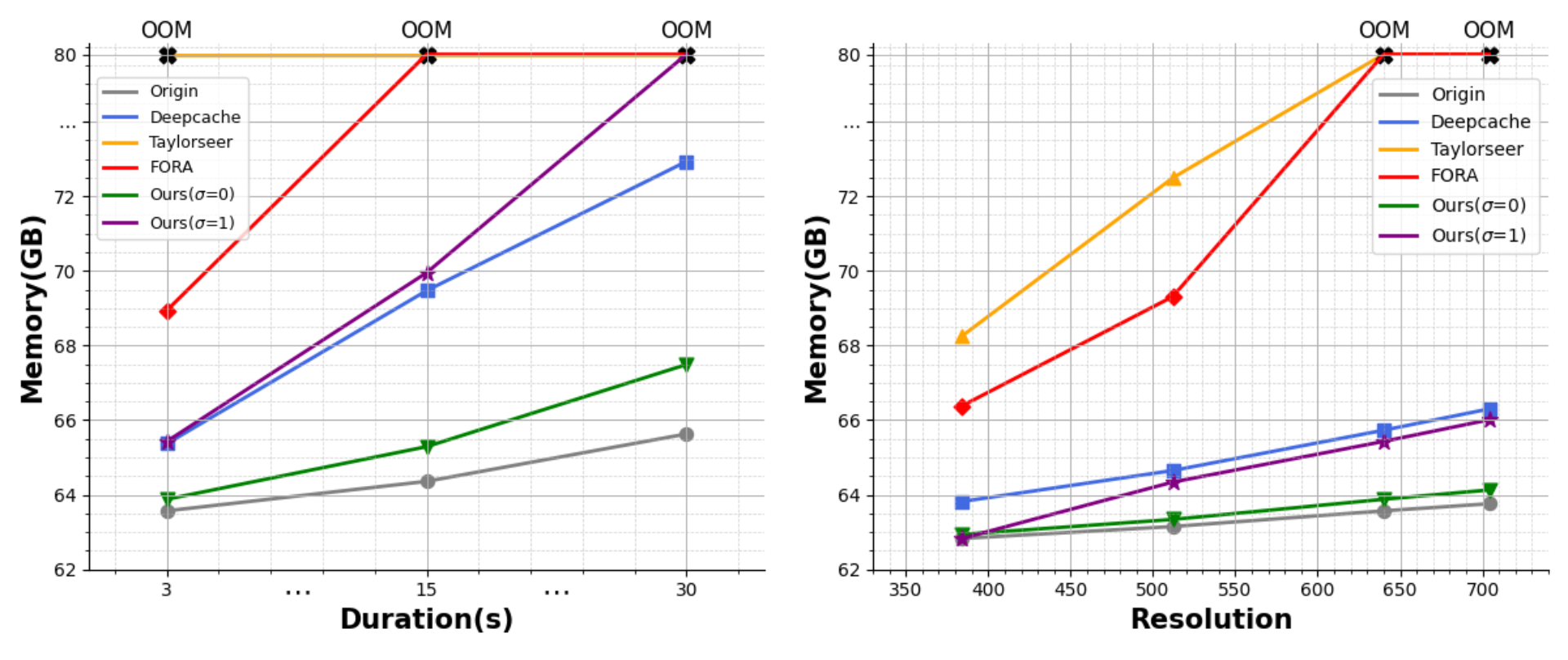}
    \caption{VRAM usage with various resolution or duration.}
    \label{fig:short-a}
  \end{subfigure}
  \hfill
  \begin{subfigure}{0.4\linewidth}
    \includegraphics[width=1\linewidth]{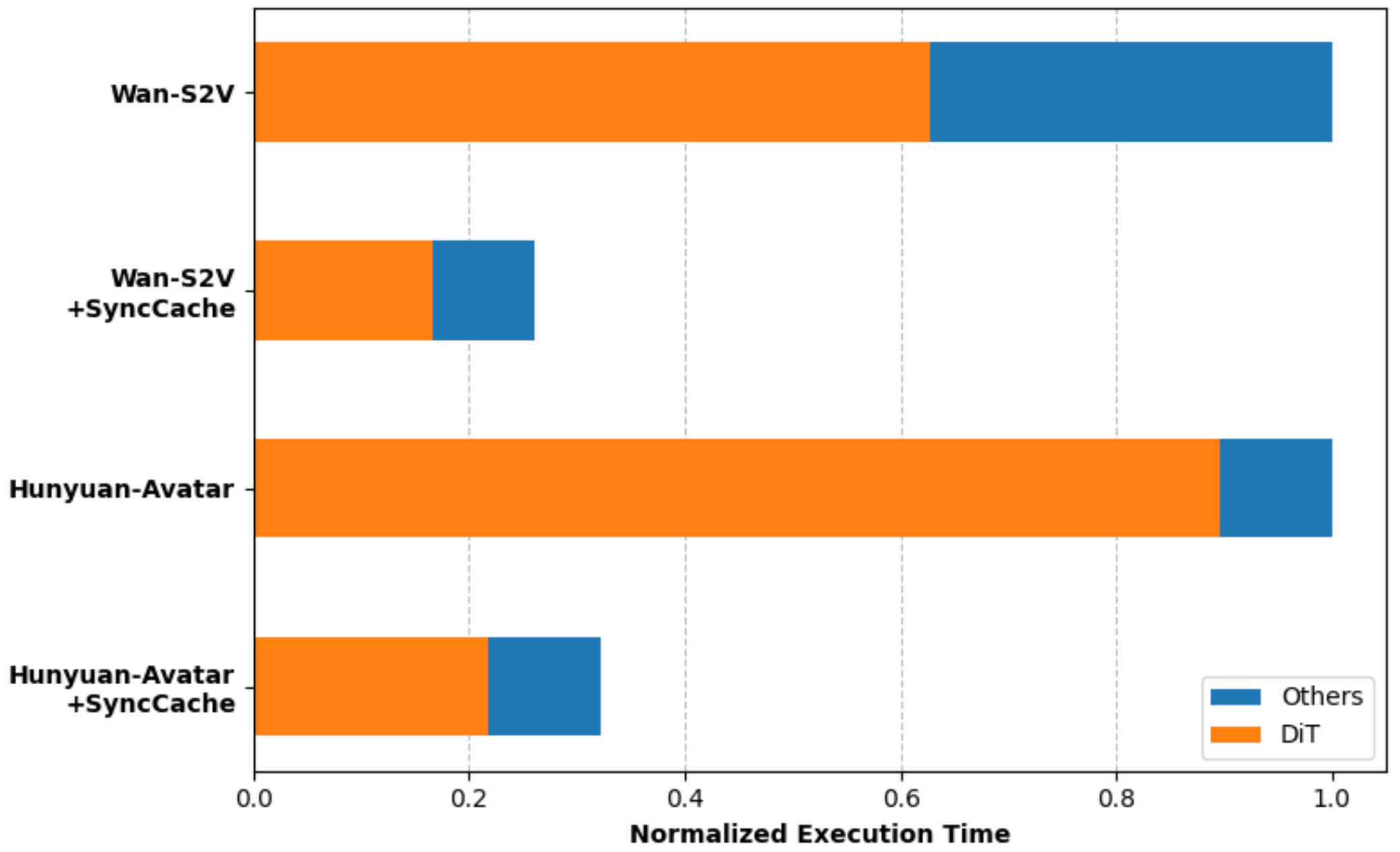}
    \caption{End-to-end latency improvement on HunyuanVideo-Avatar and Wan-S2V.}
    \label{fig:short-b}
  \end{subfigure}
  \caption{System-level Efficiency of SyncCache. (a) Unlike existing module-wise methods that rapidly exceed memory constraints as the scale of generation increases, SyncCache maintains a stable and predictable memory footprint. (b) SyncCache significantly reduces the end-to-end inference latency across diverse architectures.}
  \label{fig:memory}
\end{figure}

To accelerate diffusion inference, training-free feature caching has emerged as a prominent solution~\cite{liu2025timestep,liu2025reusing,ma2025magcache,chu2025omnicache,feng2025hicache}. However, traditional caching strategies are primarily tailored for text-to-video generation, where the text prompt serves as a low-frequency, global condition. In audio-driven portrait animation, the modality dynamics are highly asymmetric. The reference image provides a low-frequency, static visual prior for the background and identity, whereas the driving audio acts as a high-frequency, local condition that dictates rapid, frame-by-frame lip and muscle movements. Blind to this asymmetry, existing methods exhibit limitations in two distinct ways, as shown in \cref{fig:teaser}~(a). Timestep-level caching paradigms~\cite{liu2025timestep,ma2025magcache,zhou2025less} cache and skip entire denoising steps under the assumption of synchronous modality dynamics,which preserves low-frequency visual content but interrupts the continuous injection of high-frequency audio signals at high acceleration rates. As demonstrated in \cref{fig:teaser}~(c), this global skipping results in a severe loss of high-frequency details. Conversely, module-level caching~\cite{liu2025reusing,liu2025speca,zheng2025compute} attempts to preserve these details by independently caching the feature maps of all modules. This indiscriminate strategy inflates the memory footprint from $O(1)$ to $O(L)$ (where $L$ denotes the number of layers), rapidly exhausting available VRAM. As illustrated in \cref{fig:memory}~(a), increasing the target resolution and video duration causes the memory usage of TaylorSeer~\cite{liu2025reusing} and FoRA~\cite{selvaraju2024forafastforwardcachingdiffusion} to increase rapidly and exceed memory constraints.

To resolve this dilemma and accommodate asymmetric dynamics, we propose \textbf{SyncCache}, a training-free caching paradigm tailored for audio-driven portrait animation. Instead of uniform step caching or indiscriminate module caching, SyncCache explicitly exploits the dynamic asymmetry by decoupling the computation along both spatial distributions and modality streams.

Specifically, we first leverage the inherent task prior of portrait animation, in which visual dynamics are fundamentally human-centric. This characteristic naturally introduces a spatial asymmetry, where critical high-frequency variations are heavily concentrated on the human subject, whereas the background remains a low-frequency, static anchor that maintains consistency with the reference image. To effectively leverage this prior, we introduce \textit{Spatially-Asymmetric Probing}. Rather than treating the cache errors uniformly across the spatial layout, we weight the probing error with a human mask. This mechanism strategically amplifies the probing sensitivity in audio-aligned, high-frequency regions. Consequently, the model is forced to trigger timely recomputations to preserve and enhance delicate human motions.

To address modality asymmetry, we introduce \textit{Modality-Decoupled Caching}. We observe a severe computational imbalance inherent to portrait animation models: the self-attention and multilayer perceptron (MLP) layers dominate the overall computation, whereas audio blocks are lightweight but essential for injecting high-frequency signals required for precise synchronization and motion. Furthermore, our analysis reveals that the inter-block residuals bridging these audio modules exhibit temporal stability, effectively capturing the slow evolution of features across DiT blocks. Guided by this computational and dynamic asymmetry, we structurally decouple the heavy visual backbone from the audio blocks. We cache the temporally stable inter-block residuals to bypass heavy visual computations, while continuously recomputing lightweight audio blocks. This ensures the high-frequency control signals remain perfectly synchronized with negligible computational overhead.

Furthermore, to enable highly flexible deployment, SyncCache incorporates a \textit{Memory-Adaptive Optimal Selection} strategy that scales the caching memory overhead from $O(L_a)$ (where $L_a$ denotes the number of audio blocks) towards $O(1)$. We introduce a continuous cache ratio $\sigma$ to control the cache capacity. For long video generation or environments operating under strict VRAM budgets, setting a smaller $\sigma$ limits the cache to a specific subset of residual boundaries. We also discover that the temporal stability of inter-block residuals fluctuates significantly across different layers; thus, random or uniform boundary selection is inherently suboptimal and unstable. Therefore, we formulate cache selection as a dynamic programming problem to analytically identify the optimal subset of blocks that strictly minimizes inter-block temporal instability. Crucially, we observe that for any given model, this optimal subset exhibits remarkable robustness and consistency across diverse input samples. Consequently, a single, offline calibration forward pass is sufficient to map out the optimal cache plan for any $\sigma$. This zero-overhead calibration guarantees maximum generation fidelity tailored to any specific memory constraint.

Our key contributions are summarized as follows
\begin{itemize}
\item[$\bullet$] \textit{Pioneering Multimodal Caching:} We identify the fundamental limitations of existing caching methods in portrait animation and propose SyncCache, the first cache acceleration method tailored for DiT-based portrait animation that explicitly exploits asymmetric dynamics.
\item[$\bullet$] \textit{Dual-Asymmetry Decoupling:} We design \textit{Spatially-Asymmetric Probing} to amplify sensitivity in dynamic human regions, and \textit{Modality-Decoupled Caching} to continuously refresh high-frequency audio conditions, preserving visual fidelity and flawless lip-sync under extreme speedups.
\item[$\bullet$] \textit{Memory-Adaptive Caching:} We introduce a continuous cache ratio $\sigma$ to control the cache capacity to enable memory-adaptive caching. By formulating cache block selection as a dynamic programming problem through a single offline calibration, it guarantees maximum temporal stability for any given VRAM constraint without introducing online computational overhead.
\item[$\bullet$] \textit{State-of-the-Art Performance:} Extensive empirical evaluations demonstrate that SyncCache achieves a superior speed-quality trade-off. It delivers up to 4.12$\times$ acceleration on HunyuanVideo-Avatar and 3.75$\times$ on Wan-S2V while maintaining high-quality generation and audio consistency.
\end{itemize}

\section{Related Works}
\label{sec:formatting}

\subsection{Audio-driven Portrait Animation }
Portrait animation aims to synthesize a talking-face video from a reference portrait and an input audio clip~\cite{tian2024emo,gan2025omniavatar}. Early methods relied on hand-crafted intermediate representations and multi-stage pipelines \cite{wei2024aniportrait}. With the advance of diffusion models, end-to-end generation has become mainstream. Hallo~\cite{xu2024hallo} and several other methods~\cite{chen2025echomimic,meng2025echomimicv2,jiangloopy} introduce audio conditioning into U-Net-based pretrained text-to-video models by incorporating audio cross-attention, and they leverage a reference net to preserve identity consistency.  However, because U-Net-based diffusion models exhibit limited capacity, the introduction of scalable DiT~\cite{peebles2023scalable} architectures have significantly accelerated progress in the field of portrait animation. FantasyTalking~\cite{wang2025fantasytalking} and Hallo3~\cite{cui2025hallo3} explore to condition strong pretrained DiT-based video generation models on audio via cross-modality attention, whereas MultiTalk \cite{konglet} investigates audio-driven video generation in multi-speaker settings. HunyuanVideo-Avatar \cite{chen2025hunyuanvideo} is a powerful open-source model that produces dynamic, emotion-controllable, multi-character dialogue videos. Wan-S2V\cite{gao2025wan} targets film and television production and delivers realistic visuals, including natural facial expressions, coordinated body motion, and professional cinematography. Despite these advancements associated with the DiT architecture, inference speed remains a significant bottleneck. In this paper, we investigate training-free acceleration during inference and achieve substantial speedups while maintaining near-lossless fidelity.

\subsection{Diffusion Model Acceleration}
Diffusion models achieve high visual generation quality~\cite{kong2024hunyuanvideo,wan2025wan}, yet inference latency increases substantially with model capacity and architectural complexity, which hinders deployment in real applications. To address this issue, a broad set of acceleration techniques has been developed, including efficient attention~\cite{zhang2025spargeattention,xi2025svg,xia2025training}, model distillation~\cite{Wang2023VideoLCMVL,meng2023distillation}, quantization~\cite{zhaovidit,feng2025q}, improved samplers \cite{lu2022dpm}, and diffusion cache~\cite{zhang2025blockdance,chu2025omnicache,chen2025deltadit,liu2025freqca}. Training based approaches often incur additional computational overhead and require auxiliary data, limiting their practical applicability. Conversely, diffusion caching provides a training-free inference acceleration technique by exploiting feature redundancy across adjacent time steps during the sampling process of diffusion models.

Current caching strategies generally fall into two main categories. The first mainstream direction designs decision rules to cache and reuse features at the timestep level. Following this line, TeaCache~\cite{liu2025timestep} proposes a polynomial estimator to predict temporal redundancy from input differences. EasyCache~\cite{zhou2025less} replaces this polynomial function with an empirical transformation rate law. MagCache~\cite{ma2025magcache} provides a magnitude-aware strategy that adaptively skips timesteps using an error modeling mechanism. DiCache~\cite{budicache} enables the diffusion model to autonomously determine caching timings and adaptively utilize multi-step caches based on an online probe. The second group introduced by TaylorSeer~\cite{liu2025reusing} suggests combining multi-step cached features in a Taylor-expansion-like manner to predict reusable features and achieve higher speedups. ClusCa~\cite{zheng2025compute} performs spatial clustering on tokens at each timestep, computes only one token per cluster, and propagates this information to all other tokens. SpeCa~\cite{liu2025speca} introduces speculative sampling to diffusion models, predicting intermediate features for subsequent timesteps based on fully computed reference timesteps. Observing that the standard polynomial basis of a Taylor series is suboptimal for modeling the complex and non-monotonic trajectories of feature evolution in diffusion models, HiCache~\cite{feng2025hicache} proposes a Hermite polynomial-based feature cache.

\newcommand{\algcomment}[1]{%
    \Statex \hspace{-\algorithmicindent} \textcolor{gray}{\textit{#1}}%
}

\section{Method}
\label{sec:method}

\subsection{Preliminary}
\subsubsection{Flow Matching.}
Flow Matching~\cite{lipmanflow} formulates generative modeling as deterministic transport governed by an ODE that denoises samples along a straight-line path between a noise prior \(p_{\text{noise}}\) and the data distribution \(p_{\text{data}}\).
Given \(x_0 \sim p_{\text{data}}\) and \(x_T \sim p_{\text{noise}}\),  the linear interpolation at timestep \(t \in [0,T]\) is
\[
x_t \;=\; (1-\frac{t}{T})\,x_0 \;+\; \frac{t}{T}\,x_T.
\]
During training, a timestep-dependent \emph{velocity field} \(v_\theta(x_t,t,c)\) is learned to approximate the denoising direction \((x_0 - x_T)\), where \emph{c} denotes optional conditioning (e.g., text, images, or speech). This yields the neural-ODE dynamics
\[
\mathrm{d}\hat{x}_t \;=\; v_\theta(x_t,t,c)\,\mathrm{d}t.
\]
Sampling therefore reduces to integrating the learned velocity field along the straight-line trajectory from \(p_{\text{noise}}\) to \(p_{\text{data}}\).

\subsubsection{DiT-based Portrait Animation.}
Diffusion Transformers~\cite{peebles2023scalable} have become the dominant backbone for video generation. In portrait animation, the model typically builds upon a pretrained DiT-based image-to-video backbone with interleaving additional audio blocks. These lightweight audio blocks act as high-frequency local controllers, injecting precise lip-sync and human motion signals into the massive backbone.

\begin{figure}[tb]   %
  \centering
  \includegraphics[width=1\textwidth]{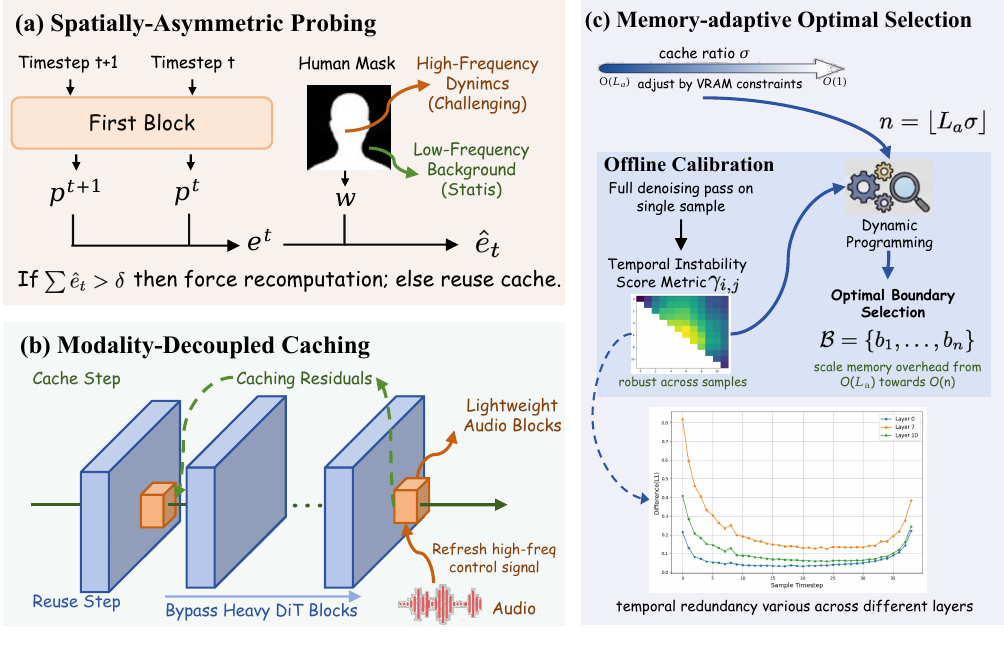}  
  \caption{\textbf{Overview of the SyncCache framework.} SyncCache accelerates audio-driven portrait animation by explicitly exploiting inherent asymmetric dynamics. \textit{(a)} We utilizes a human mask to prioritize computations in highly dynamic human regions. \textit{(b)} We physically isolates the conditioning stream, bypassing heavy visual DiT blocks via cached residuals while continuously refreshing lightweight audio blocks to maintain flawless lip-sync. (c) We employ a offline calibration and Dynamic Programming (DP) to analytically determine the optimal caching boundaries, gracefully scaling the VRAM footprint according to the cache ratio $\sigma$.}
  \label{fig:method}
\end{figure}

\subsection{SyncCache}

To elegantly accommodate the complex dynamics of audio-driven portrait animation, we propose SyncCache, a training-free acceleration paradigm with dual-ssymmetry decoupling. Instead of uniform step caching or indiscriminate module caching, SyncCache explicitly exploits the dynamic asymmetry inherent to the task by decoupling the computation along both spatial distributions and modality streams. The overall framework is illustrated in \cref{fig:method}.
\begin{figure}[tb]
  \centering
  \includegraphics[width=1\textwidth]{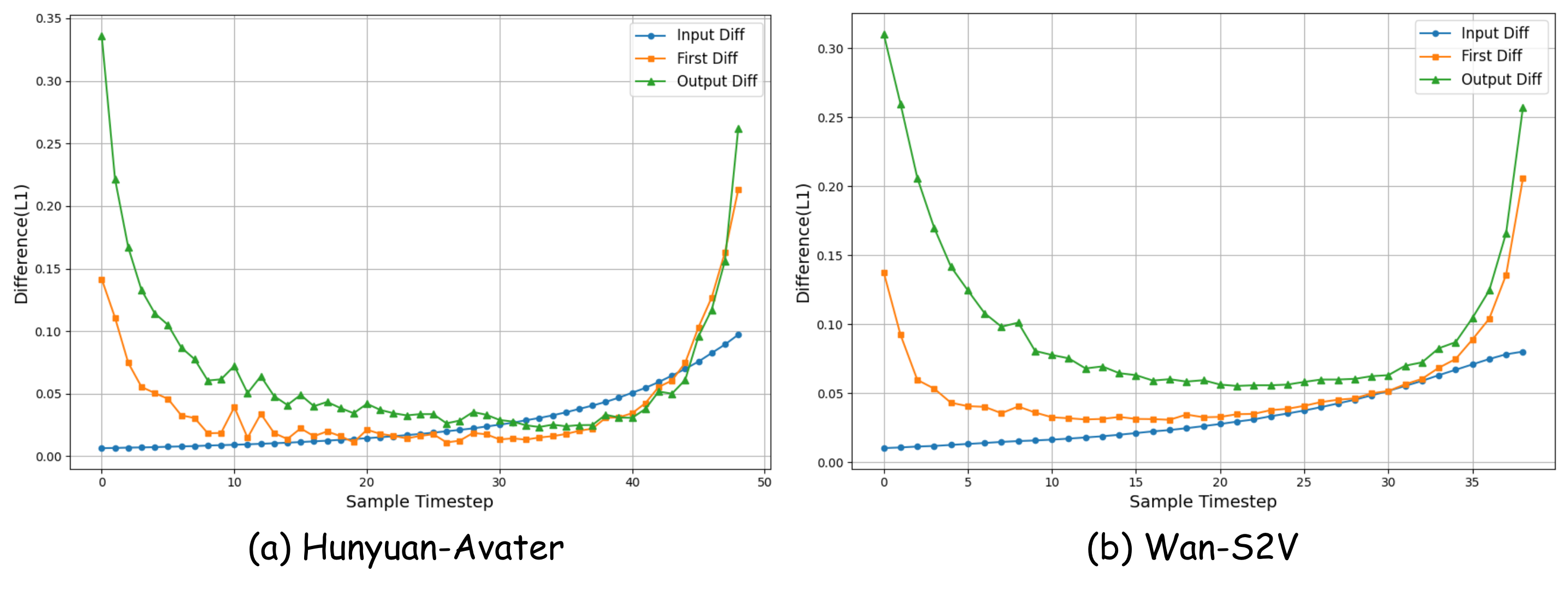}  
  \caption{\textbf{Visualization of input differences and output differences
in consecutive timesteps of HunyuanVideo-Avatar and Wan-S2V.} The
difference of the first transformer block output exhibits a strong
correlation with the difference of the final output between adjacent
timesteps in portrait animation models}
  \label{fig:visual}
\end{figure}
\paragraph{Spatially-Asymmetric Probing.} 
To accelerate inference via feature caching, it is crucial to estimate the temporal redundancy at each denoising step and determine when to cache. Motivated by prior studies~\cite{guan2025forecasting,liu2025timestep}, we observe that the feature variation of the first block strongly correlates with the overall denoising dynamics across adjacent timesteps, as shown in \cref{fig:visual}. Therefore, we employ the output of the first block, denoted as $p^t$, as a lightweight probe to dynamically determine when to execute a full forward pass and when to safely reuse cache. At each step \(t\), we compute the first block output \(p^{\,t}\,\) and measure the relative \(L1\) error with respect to the preceding step as a proxy. 
\[
e_t = \mathrm{L1}_{\mathrm{rel}}(p^{\,t}\,,p^{\,t+1}\,) = \frac{\left\|\, p^{\,t}\, - p^{\,t+1}\, \,\right\|_{1}} {\left\|\, p^{\,t+1}\, \,\right\|_{1}} \, .
\]
However, directly relying on a global probing metric overlooks a fundamental task prior: portrait animation exhibits severe spatial asymmetry. Critical high-frequency variations are intensely localized on the human subject, whereas the background remains a low-frequency, static anchor that maintains consistency with the reference image. To strictly enforce the sensitivity of the model to these critical high-frequency details, we modulate the probing error using a spatial human mask $M$. Crucially, the acquisition of the mask entails minimal computational overhead. It is inherently available as a standard auxiliary input in many DiT-based models~\cite{cui2025hallo3,chen2025hunyuanvideo}, and for mask-free architectures~\cite{wan2025wan}, $M$ is seamlessly extracted via a lightweight human detector with sub-second latency. The spatially-asymmetric error $\hat{e}_t$ at timestep $t$ is thus formulated as:

\[
\hat e_t \;=\frac{\left\|\,( p^{\,t}\, - p^{\,t+1})\odot \bigl(1 + \omega M\bigr)\, \,\right\|_{1}} {\left\|\, p^{\,t+1}\, \,\odot \bigl(1 + \omega M\bigr)\right\|_{1}} \,  ,
\]
where $\omega$ is a human emphasis weight that amplifies probing sensitivity in human regions, and $\odot$ denotes element-wise multiplication.

By strategically tracking this modulated error, we accumulate $\hat{e}_t$ during the cached steps. Once the accumulated error exceeds a user-specified tolerance threshold $\delta$ ($\sum \hat{e}_t > \delta$), the model is forced a full forward pass to refresh the cache. This ensures the model instantly triggers recomputations to preserve delicate human motions and triggers timely recomputation to correct error.

\paragraph{Modality-Decoupled Caching.}
To tackle the inherent modality asymmetry in audio-driven portrait animation, SyncCache introduces a structural separation of the computationally intensive backbone from audio blocks. Within the DiT-based portrait animation pipeline, audio blocks are lightweight and contribute less than 1\% to the end-to-end latency, whereas self attention and MLP layers dominate the cost. Despite this low computational overhead, audio blocks are essential for portrait animation and continuously injecting the high-frequency signals required for precise lip synchronization and human motion. Furthermore, as illustrated in \cref{fig:method}~(c), our empirical analysis reveals that the inter-block residuals between these audio modules exhibit high temporal stability across consecutive denoising steps, making them ideal candidates for feature caching.

Guided by this asymmetry, during a timestep with full computation $t_a$, we cache these stable inter-block residuals $r_{i,j}^{t_a}$ between the $i$-th and $j$-th audio block. During subsequent reuse steps, we bypass the computationally expensive DiT blocks by directly reusing the cached residuals $r_{i,j}^{t_a}$. Concurrently, we still compute the lightweight audio blocks. This decoupling ensures that the high-frequency control signals are updated and synchronized at every timestep.
\subsection{Memory-adaptive Optimal Selection}
\label{alg_method}

Although \textit{Modality-Decoupled Caching} successfully accelerates inference, caching the inter-block residuals at every audio block boundary still scales the memory footprint to $O(L_a)$, where $L_a$ is the total number of audio blocks. For long video generation or deployment on VRAM-constrained devices, this accumulation can rapidly exhaust available memory, as shown in~\cref{fig:memory}~(a). To enable flexible deployment across diverse hardware constraints, it is essential to dynamically adapt the cache capacity to the specific VRAM budget. Therefore, we introduce a continuous cache ratio $\sigma \in (0, 1]$ that controls the cache capacity to a specific subset of $n = \lfloor L_a \sigma \rfloor$ residual boundaries. However, as shown in \cref{fig:method}~(c), our empirical analysis reveals that the temporal stability of these inter-block residuals varies across different network depths. Consequently, uniform or random selection which boundaries to cache is inherently suboptimal and unstable.

To minimize the global inter-block temporal instability, we formulate this boundary selection as a dynamic programming (DP) problem. We quantify the temporal instability score $\gamma_{i,j}$ for the residual segment between the $i$-th and $j$-th audio blocks over the full denoising trajectory $T$ as:
\begin{equation}\gamma_{i,j} = \sum_{t=0}^{T-1} \frac{||r_{i,j}^t - r_{i,j}^{t+1}||_1}{||r_{i,j}^{t+1}||_1}
\end{equation}

Crucially, we observe that for any given model, the optimal path $\mathcal{B}$ exhibits strong robustness across diverse input audio and reference samples. Consequently, a single offline calibration pass is sufficient to determine the optimal cache plan for any specified $\sigma$. This guarantees maximum generation fidelity tailored to the VRAM budget, while eliminating online computational overhead.

\begin{algorithm}[t]
\caption{Memory-adaptive Caching}
\label{alg}
\begin{algorithmic}[1]
\State \textbf{Inputs} number of audio blocks \(N\) and cache ratio \(\sigma\)
\State \textbf{Output} optimal cache boundaries \(\mathcal{B}=\{b_1,\dots,b_n\}\) with \(0=b_0<b_1<\dots<b_n=N\)

\State \textbf{Calibration} run one full denoising pass on a sample and compute \(\gamma_{i,j}\) for all \(0\le i<j\le N\)
\[
\gamma_{i,j}=\sum_{t=0}^{T-1}\frac{\left\|r_{i,j}^{\,t}-r_{i,j}^{\,t+1}\right\|_{1}}{\left\|r_{i,j}^{\,t+1}\right\|_{1}}
\]

\State \(n \gets \lfloor N\sigma \rfloor\)

\State \(\mathcal{B} \gets \textsc{CacheSearch}\big(\{\gamma_{i,j}\}, n\big)\) 
\Statex \textit{Execute dynamic programming to select \(n\) optimal cache boundaries}

\State \textbf{Return} \(\mathcal{B}\)

\end{algorithmic}
\end{algorithm}

\section{ Experiments}
\subsection{Experimental Settings}
We conduct our main experiments on two portrait animation models, Wan-S2V~\cite{gao2025wan} and HunyuanVideo-Avatar~\cite{chen2025hunyuanvideo}, to evaluate the effectiveness of our method. HunyuanVideo-Avatar is a portrait animation model built on HunyuanVideo, which injects audio conditioning by inserting audio blocks after the dual-stream DiT blocks. We follow the standard 50-step inference protocol as the baseline and keep all default sampling parameters to ensure strict experimental consistency. Wan-S2V is an audio-driven video generation model built on Wan-14B. It injects audio conditioning by inserting eleven audio blocks after backbone blocks and it uses a 40-step UniPC sampler.
\paragraph{Evaluation and Metrics}
Following prior work \cite{gao2025wan,konglet,li2025infinityhuman} in portrait animation, we perform evaluations on the EMTD dataset \cite{meng2025echomimicv2}, which primarily consists of solo-talking and semi-body human videos. To assess the quality of the talking-face videos generated under acceleration, we evaluate three aspects: inference efficiency, visual quality, and audio-visual synchronization. Regarding inference efficiency, we report the speedup ratio and the inference latency. For visual quality, in line with previous studies on caching acceleration \cite{liu2025timestep,chen2025deltadit,liu2025reusing}, we utilize LPIPS \cite{zhang2018lpips}, PSNR \cite{hore2010psnr}, and SSIM \cite{wang2004ssim} to evaluate the fidelity of generated videos relative to origin results. Consistent with the portrait animation methods\cite{xu2024hallo,chen2025echomimic,tian2024emo}, we additionally employ Frechet Inception Distance (FID) \cite{heusel2017fid} and Frechet Video Distance (FVD) \cite{unterthiner2019fvd} to measure the distance between generated and real videos. To evaluate the alignment between lip movements and audio signals, we use Sync-C and Sync-D \cite{chung2016syncnet}, metrics that are widely adopted within the portrait animation community.

\paragraph{Implementation Details}
All experiments are conducted on NVIDIA A800 80GB GPUs with PyTorch and FlashAttention \cite{dao2022flashattention} enabled by default. For latency benchmarking, we use 
8$\times$ A800 GPUs. We apply the default FSDP configuration to HunyuanVideo-Avatar. For Wan-S2V, we use the default FSDP to shard the DiT and the text encoder together with Ulysses sequence parallelism. We set the human emphasis weight  \(\omega\) to 2 (empirically robust for $\omega \in [2, 4]$). For mask-free models (e.g., Wan-S2V), we employ the same lightweight detection model (containing 46 million parameters) as utilized in Hunyuan-Avatar, which incurs a negligible latency of 0.07 seconds. 

\subsection{Quantitative Comparison}
\label{sec:quti}
\begin{table}[tb]
	\centering
	\caption{\textbf{ Quantitative comparison with other methods on HunyuanVideo-Avatar.} The best value is in \textbf{bold} and the second best is \underline{underlined}. \textbf{OOM} indicates a CUDA out-of-memory error on eight A800 GPUs with 80GB memory each. \textsuperscript{\dag}TaylorSeer-series denotes a set of recent methods derived from TaylorSeer.}
    \label{hunyuan}
	 \resizebox{1\linewidth}{!}{
     \begin{tabular}{c|ccccc|cc|cc}
    \toprule
    \multicolumn{1}{c|}{\multirow{2}*{\textbf{Method}}} & \multicolumn{5}{c|}{\multirow{1}*{\textbf{Visual Quality}}} & \multicolumn{2}{c|}{\multirow{1}*{\textbf{Audio Consistency}}} & \multicolumn{2}{c}{\multirow{1}*{\textbf{Acceleration}}} \\
    \cmidrule(lr){2-6}
    \cmidrule(lr){7-8}
    \cmidrule(lr){9-10}
     & \textbf{LPIPS$\downarrow$} & \textbf{PSNR$\uparrow$} & \textbf{SSIM$\uparrow$} & \textbf{FID$\downarrow$} & \textbf{FVD$\downarrow$} & \textbf{Sync-C$\uparrow$} & \textbf{Sync-D$\downarrow$} & \textbf{Speedup$\uparrow$} & \textbf{Latency(s)$\downarrow$} \\
    \midrule 
    \text{Original: 50 steps } & - & - & - & 25.27 & 240.16 & 6.963 & 8.640 &  - & 524\\
        \text{$\Delta$-DiT \cite{chen2025deltadit}} & 0.1321 & 23.26 & 0.8281 &  26.52 & 251.68 & 6.652 & 8.878 & 1.38$\times$ & 381 \\
        \text{TeaCache \cite{liu2025timestep}} & 0.1730 & \underline{25.55} & 0.8428 & 26.83 & 236.08 & 6.842 & 8.712 & 2.25$\times$ & 233\\
            \text{MagCache\cite{ma2025magcache}} & 0.1696 & \textbf{25.76} & 0.8455 & 26.43 & \underline{235.87} & 6.830 & \underline{8.664} & 2.30$\times$ & 228 \\

    \text{DiCache~\cite{budicache}} & 0.1548 & 25.29 & 0.8490 & \underline{26.12} & 238.15 & 6.834 & 8.726 & 2.41$\times$ & 217 \\
    \text{TaylorSeer-series\textsuperscript{\dag}} \cite{liu2025reusing,liu2025speca,zheng2025compute} & OOM & OOM & OOM & OOM & OOM & OOM & OOM & - & - \\
    \text{CGCache \cite{guan2025forecasting}} & 0.1848 & 24.05 & 0.8249 & 27.75 & 238.51 &  6.814 & 8.755 & 3.18$\times$ & 164 \\
    \textbf{SyncCache-slow} & \textbf{0.1016} & 24.93 & \textbf{0.8618} &\textbf{25.65} & \textbf{234.86} & \textbf{6.944} &\textbf{ 8.653} &  \underline{3.34$\times$} & \underline{157}\\
    \textbf{SyncCache-fast}& \underline{0.1172} & 24.41 & \underline{0.8493} &26.89 & 241.27 & \underline{6.902} & 8.673 &  \textbf{4.12$\times$} & \textbf{127}\\
    \bottomrule
	\end{tabular}}

\end{table}
\subsubsection{Performance on HunyuanVideo-Avatar.}
As reported in~\cref{hunyuan}, SyncCache delivers the best overall performance, surpassing existing baselines across almost all key metrics. Under a conservative threshold, SyncCache-slow reduces end-to-end latency from nearly 10 minutes to 157 seconds while achieving the best visual fidelity across LPIPS (0.1016), SSIM (0.8618), FID (25.65), and FVD (234.86). Most crucially, regarding audio consistency, SyncCache-slow achieves a near-lossless Sync-C of 6.944 (vs. 6.963 Original), markedly higher than existing methods like TeaCache and MagCache. This validates that our strategy of exploiting asymmetric dynamic is essential for preserving delicate high-frequency human dynamics. Pushing to a $4.12\times$ speedup, SyncCache-fast still dominate all baselines in both perceptual quality and lip-sync alignment. In contrast, existing methods reveal intrinsic limitations due to their blind uniform assumption. Timestep-level caching (TeaCache, MagCache) saturates speedups around $2.30\times$ and discards high-frequency audio conditions, leading to a noticeable degradation in Sync-C. TaylorSeer-series exceed memory constraint and trigger CUDA out-of-memory (OOM) failures due to indiscriminate $O(L)$ feature caching.
\begin{table}[tb]
	\centering
	\caption{\textbf{Quantitative comparison with other methods on Wan-S2V}. The best value is in \textbf{bold} and the second best is \underline{underlined}. \textbf{OOM} indicates a CUDA out-of-memory error on eight A800 GPUs with 80GB memory each. \textsuperscript{\dag}TaylorSeer-series denotes a set of recent methods derived from TaylorSeer.}
    \label{wan_supp}
	\resizebox{1\linewidth}{!}{
    \begin{tabular}{c|ccccc|cc|cc}
    \toprule
    \multicolumn{1}{c|}{\multirow{2}*{\textbf{Method}}} & \multicolumn{5}{c|}{\multirow{1}*{\textbf{Visual Quality}}} & \multicolumn{2}{c|}{\multirow{1}*{\textbf{Audio Consistency}}} & \multicolumn{2}{c}{\multirow{1}*{\textbf{Acceleration}}} \\
    \cmidrule(lr){2-6}
    \cmidrule(lr){7-8}
    \cmidrule(lr){9-10}
     & \textbf{LPIPS$\downarrow$} & \textbf{PSNR$\uparrow$} & \textbf{SSIM$\uparrow$} & \textbf{FID$\downarrow$} & \textbf{FVD$\downarrow$} & \textbf{Sync-C$\uparrow$} & \textbf{Sync-D$\downarrow$} & \textbf{Speedup$\uparrow$} & \textbf{Latency (s)$\downarrow$} \\
    \midrule 
Original (40 steps) & -- & -- & -- & 36.60 & 293.05 & 6.712 & 8.632 & -- & 113 \\
    TeaCache~\cite{liu2025timestep} & 0.1863 & 19.04 & 0.7468 & 39.45 & 297.45 & 6.678 & 8.642 & 2.93$\times$ & 38.62 \\
    MagCache~\cite{ma2025magcache} & 0.1839 & \underline{19.46} & \underline{0.7524} & \textbf{32.93 }& 287.36 & 6.709 & 8.641 & 2.96$\times$ & 38.15 \\
    $\Delta$-DiT~\cite{chen2025deltadit} & 0.1869 & 19.24 & 0.7445 & 38.91 & \textbf{277.33} & 6.693 & 8.616 & 1.73$\times$ & 65.26 \\
    CGCache~\cite{guan2025forecasting} & 0.1884 & 19.20 & 0.7443 & 37.17 & 298.76 & 6.710 & 8.651 & 3.00$\times$ & 37.65 \\
    Taylorseer-series\textsuperscript{\dag}~\cite{zheng2025compute,liu2025reusing,liu2025speca} & OOM & OOM & OOM & OOM & OOM & OOM & OOM & - & - \\
    \text{DiCache~\cite{budicache}} & \underline{0.1835} & 19.43 & 0.7498 & 34.17 & 281.56 & \underline{6.716 }& \underline{8.614} & 2.99$\times$ & 37.82 \\
    \textbf{SyncCache} & \textbf{0.1775} & \textbf{19.80} & \textbf{0.7665} & \underline{33.83} & \underline{280.78} & \textbf{6.791} & \textbf{8.541} & \textbf{3.75$\times$} & \textbf{30.15} \\

    \bottomrule
	\end{tabular}}
\end{table}
\subsubsection{Performance on Wan-S2V.}
Wan-S2V presents a more severe challenge due to its extremely compressed 40-step sampler and complex dynamic patterns. However, as demonstrated in~\cref{wan_supp}, SyncCache effectively adapts to this challenging context, demonstrating substantial improvements over existing baselines. When traditional caching methods are applied to Wan-S2V, the lack of explicit modality modeling leads to compromised performance. Both TeaCache and MagCache exhibit observable degradations in audio consistency and visual fidelity. The step-wide skipping mechanism of these methods interrupts the continuous high-frequency audio feed required to stabilize visual motion, which results in noticeable lip-sync misalignment. In contrast, SyncCache successfully addresses this issue by  targeting the dual asymmetry in the dynamics of portrait animation. As a result, SyncCache dominates across all quality metrics on Wan-S2V, attaining the best LPIPS (0.1775), PSNR (19.80), and Sync-C (6.791) at $3.75\times$ speedups. This consistent success generalizes the necessity of exploiting asymmetric dynamics to preserve delicate high-frequency motion.

\subsection{Qualitative Comparison}
\begin{figure}[tb]
  \centering
\includegraphics[width=1.0\linewidth]{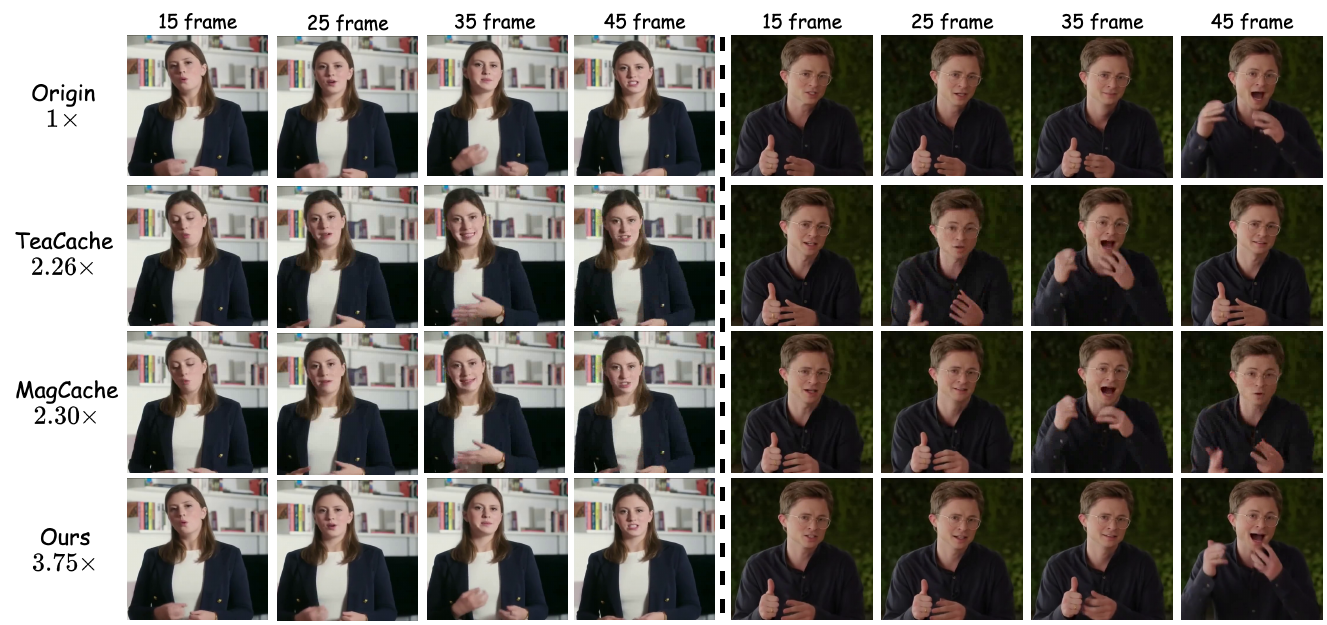}
  \caption{Comparison of visual quality and efficiency (denoted by speedup ratio) with other methods on HunyuanVideo-Avatar.}
  \label{fig:vis}
\end{figure}

\begin{figure}[tb]
    \centering
    \includegraphics[width=1\linewidth]{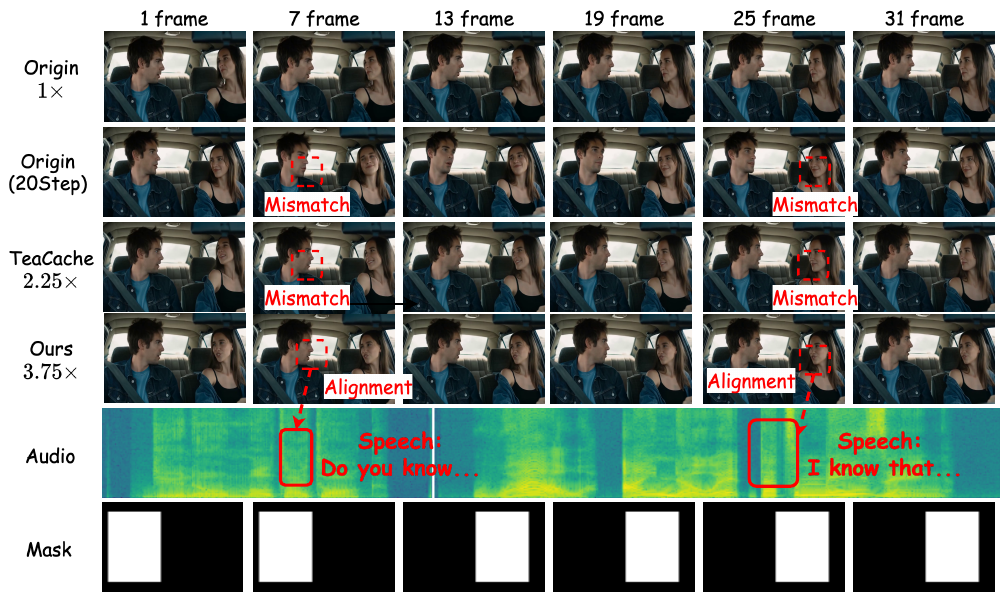}
    \caption{Qualitative comparison in multi-person scenarios. Compared with the original method and TeaCache~\cite{liu2025timestep}, our method exhibits stronger consistency for the utterances ``Do you know...'' and ``I know that...'' in multi-speaker scenarios.}
    \label{fig:sup_vis}
\end{figure}
We provide visualizations to compare the quality of generated videos against that of several baseline methods. As demonstrated in \cref{fig:vis}, our method better preserves high-frequency details and temporal alignment at higher speedup ratios. Traditional caching methods, such as MagCache and TeaCache, operate under synchronous modality dynamics, which leads to the loss of high-frequency details. This degradation manifests as blurry hand gestures and poor lip-synchronization. In contrast, SyncCache achieves notably better lip-audio alignment compared to other approaches at an acceleration factor of 4.12$\times$. 
\subsubsection{Effectiveness in Multi-Person Scenarios.}
Research on portrait animation is actively expanding to multi-speaker scenarios, often employing a dynamic masking strategy to control the range of audio cross attention constraints~\cite{konglet,chen2025hunyuanvideo}. We integrate this masking strategy into HunyuanVideo-Avatar and evaluate the compatibility of the proposed method with this approach. As shown in \cref{fig:sup_vis}, SyncCache produces significantly finer and more accurate lip synchronization details in multi-speaker dialogue scenarios.  We attribute this improvement to Modality-Decoupled Caching (MDC). Baseline methods such as TeaCache experience synchronization failures due to the weak control of audio conditions, resulting in misaligned lip movements. In contrast, MDC ensures that the audio condition is reinforced at every step. By continuously refreshing the audio controls and incorporating mask information, MDC enables SyncCache to operate seamlessly with complex spatial masks without sacrificing temporal alignment.

\subsection{Ablation Study}
\begin{table}[tb]
	\centering
	\caption{Ablation study of SyncCache components on HunyuanVideo-Avatar. }
    \label{abaltion1}
    \small
     \begin{tabular}{c|ccc|cc|c}
    \toprule
    Method& LPIPS$\downarrow$ & PSNR$\uparrow$ & SSIM$\uparrow$ &  Sync-C$\uparrow$ & Sync-D$\downarrow$ &  Latency(s)$\downarrow$ \\
    \midrule 
    w/o SAP & 0.1259 & 24.76 & 0.8534 & 6.867 & 8.684  & 161\\
     w/o MDC & 0.1571 & 24.54 & 0.8437 & 6.822 & 8.735  & \textbf{156}\\
      \textbf{Ours} & \textbf{0.1016} & \textbf{24.93} & \textbf{0.8618} & \textbf{6.944} &\textbf{ 8.653}  & 157\\
    \bottomrule
	\end{tabular}

\end{table}
\subsubsection{Ablation of Core SyncCache Components.}
To validate the effectiveness of the proposed design, we examine the contributions of Spatially-Asymmetric Probing (SAP) and Modality-Decoupled Caching (MDC).
\paragraph{Effect of Spatially-Asymmetric Probing.} Removing SAP (i.e., reverting to a naive, uniform probing metric) leads to a clear degradation in spatial visual fidelity. As demonstrated in \cref{abaltion1}, all metrics noticeably deteriorates without SAP. This empirical drop confirms the initial intuition: without explicitly prioritizing highly dynamic human regions, the probing mechanism suffers from global dilution. Consequently, it prematurely skips computations and fails to preserve intricate, high-frequency human details. In cases where the confidence of the human mask is low or the detection fails, our pipeline naturally defaults to using the full image. Crucially, Spatially-Asymmetric Probing utilizes a soft multiplicative mask rather than a hard crop. Therefore, an imperfect mask simply causes the model to degrade gracefully toward uniform probing, rather than imposing incorrect spatial asymmetries.

\paragraph{Effect of Modality-Decoupled Caching.} Disabling MDC equates to a rigid, synchronous caching strategy where both visual and audio blocks are skipped simultaneously. This omission causes a significant degradation in temporal alignment, with the Sync-C metric decreasing from 6.944 to 6.822. Crucially, a comparison between the configuration without MDC and the proposed approach reveals that restoring the continuous recomputation of audio blocks recovers near-lossless lip-sync accuracy with a slight latency penalty. This finding demonstrates that isolating and continuously refreshing the lightweight audio stream is a necessity with zero overhead for preserving high-frequency synchronization.
\begin{table}[tb]
	\centering
	\caption{Ablation study of Memory-adaptive Optimal Selection on HunyuanVideo-Avatar with $\sigma=0.4$. }
    \label{abaltion2}
    \small
     \begin{tabular}{c|ccc|cc}
    \toprule
    Method& LPIPS$\downarrow$ & PSNR$\uparrow$ & SSIM$\uparrow$ &  Sync-C$\uparrow$ & Sync-D$\downarrow$  \\
    \midrule 
      1 Sample & 0.1016 & 24.93 & \textbf{0.8618} & 6.944 & 8.653\\
          All Samples & \textbf{0.0981} & \textbf{25.01} & 0.8603 & \textbf{6.951} & \textbf{8.645} \\
         Silent Sample & 0.1021 & 24.94 & 0.8608 & 6.932 & 8.671\\
         w/o DP & 0.1369 & 24.69 & 0.8526 & 6.847 & 8.709\\
    \bottomrule
	\end{tabular}
\end{table}
\subsubsection{Effectiveness and Robustness of Memory-adaptive Optimal Selection.}
We evaluate the Memory-adaptive Optimal Selection strategy regulated by dynamic programming (DP) to verify the effectiveness and robustness of the method across diverse samples. As demonstrated in \cref{abaltion2}, analytically identifying the optimal subset of boundaries based on residual stability achieves superior generalization compared to arbitrary selection. Compared to the arbitrary selection setting (w/o DP), the proposed DP-based strategy improves performance, reducing the LPIPS from 0.1369 to 0.1016 and increasing the Sync-C from 6.847 to 6.944. Furthermore, the robustness and consistency of this optimal path remain stable across different calibration scenarios. We observe negligible performance differences when optimizing the cache path based on a single sample instead of the average of all samples. Even utilizing a silent sample with minimal lip motion to determine the cache boundaries yields results comparable to those obtained from active talking samples. This result indicates that a single offline calibration pass can satisfy specific memory constraints without introducing online computational overhead. These findings confirm that optimal cache boundaries depend on the inter-layer temporal dynamics of the model rather than specific input prompts, thereby validating the robustness of the proposed calibration method.

\section{Conclusion}
We propose SyncCache, a training-free, memory-adaptive caching strategy tailored for DiT-based audio-driven portrait animation. By explicitly exploiting the asymmetric dynamics of the generation process, we developed Spatially-Asymmetric Probing to prioritize high-frequency human motion regions and Modality-Decoupled Caching to continuously refresh lightweight audio conditions while bypassing heavy, redundant visual computations. Furthermore, we incorporated a memory-adaptive optimal selection strategy that leverages offline dynamic programming to dynamically scale cache capacity according to VRAM constraints without incurring online overhead.
Experiments show that SyncCache achieves a superior speed–quality trade-off, delivering a 4.12× acceleration on HunyuanVideo-avatar and 3.75× on Wan-S2V with negligible degradation in visual quality or audio alignment.

\section*{Acknowledgements}
This work was supported in part by the Natural Science Foundation of China (No. 62332002, 62425101), The Guangdong Grants (Grant No.2023ZT10X075), and Shenzhen KQTD (No.20240729102051063).


%
%
\bibliographystyle{splncs04}
\bibliography{main}

@String(AAAI = {AAAI})

@inproceedings{chen2025echomimic,
  title={Echomimic: Lifelike audio-driven portrait animations through editable landmark conditions},
  author={Chen, Zhiyuan and Cao, Jiajiong and Chen, Zhiquan and Li, Yuming and Ma, Chenguang},
  booktitle={Proceedings of the AAAI Conference on Artificial Intelligence},
  volume={39},
  number={3},
  pages={2403--2410},
  year={2025}
}

@inproceedings{meng2025echomimicv2,
  title={Echomimicv2: Towards striking, simplified, and semi-body human animation},
  author={Meng, Rang and Zhang, Xingyu and Li, Yuming and Ma, Chenguang},
  booktitle={Proceedings of the Computer Vision and Pattern Recognition Conference},
  pages={5489--5498},
  year={2025}
}

@article{xu2024hallo,
  title={Hallo: Hierarchical audio-driven visual synthesis for portrait image animation},
  author={Xu, Mingwang and Li, Hui and Su, Qingkun and Shang, Hanlin and Zhang, Liwei and Liu, Ce and Wang, Jingdong and Yao, Yao and Zhu, Siyu},
  journal={arXiv preprint arXiv:2406.08801},
  year={2024}
}

@inproceedings{cui2025hallo3,
  title={Hallo3: Highly dynamic and realistic portrait image animation with video diffusion transformer},
  author={Cui, Jiahao and Li, Hui and Zhan, Yun and Shang, Hanlin and Cheng, Kaihui and Ma, Yuqi and Mu, Shan and Zhou, Hang and Wang, Jingdong and Zhu, Siyu},
  booktitle={Proceedings of the Computer Vision and Pattern Recognition Conference},
  pages={21086--21095},
  year={2025}
}

@inproceedings{peebles2023scalable,
  title={Scalable diffusion models with transformers},
  author={Peebles, William and Xie, Saining},
  booktitle={Proceedings of the IEEE/CVF international conference on computer vision},
  pages={4195--4205},
  year={2023}
}

@article{zhou2025less,
  title={Less is Enough: Training-Free Video Diffusion Acceleration via Runtime-Adaptive Caching},
  author={Zhou, Xin and Liang, Dingkang and Chen, Kaijin and Feng, Tianrui and Chen, Xiwu and Lin, Hongkai and Ding, Yikang and Tan, Feiyang and Zhao, Hengshuang and Bai, Xiang},
  journal={arXiv preprint arXiv:2507.02860},
  year={2025}
}

@inproceedings{jiangloopy,
  title={Loopy: Taming Audio-Driven Portrait Avatar with Long-Term Motion Dependency},
  author={Jiang, Jianwen and Liang, Chao and Yang, Jiaqi and Lin, Gaojie and Zhong, Tianyun and Zheng, Yanbo},
  booktitle={The Thirteenth International Conference on Learning Representations},
  year={2025},
}

@inproceedings{tian2024emo,
  title={Emo: Emote portrait alive generating expressive portrait videos with audio2video diffusion model under weak conditions},
  author={Tian, Linrui and Wang, Qi and Zhang, Bang and Bo, Liefeng},
  booktitle={European Conference on Computer Vision},
  pages={244--260},
  year={2024},
  organization={Springer}
}

@article{gan2025omniavatar,
  title={OmniAvatar: Efficient Audio-Driven Avatar Video Generation with Adaptive Body Animation},
  author={Gan, Qijun and Yang, Ruizi and Zhu, Jianke and Xue, Shaofei and Hoi, Steven},
  journal={arXiv preprint arXiv:2506.18866},
  year={2025}
}

@article{gao2025wan,
  title={Wan-s2v: Audio-driven cinematic video generation},
  author={Gao, Xin and Hu, Li and Hu, Siqi and Huang, Mingyang and Ji, Chaonan and Meng, Dechao and Qi, Jinwei and Qiao, Penchong and Shen, Zhen and Song, Yafei and others},
  journal={arXiv preprint arXiv:2508.18621},
  year={2025}
}

@article{xi2025svg,
  title={Sparse videogen: Accelerating video diffusion transformers with spatial-temporal sparsity},
  author={Xi, Haocheng and Yang, Shuo and Zhao, Yilong and Xu, Chenfeng and Li, Muyang and Li, Xiuyu and Lin, Yujun and Cai, Han and Zhang, Jintao and Li, Dacheng and others},
  journal={arXiv preprint arXiv:2502.01776},
  year={2025}
}

@inproceedings{xia2025training,
  title={Training-free and adaptive sparse attention for efficient long video generation},
  author={Xia, Yifei and Ling, Suhan and Fu, Fangcheng and Wang, Yujie and Li, Huixia and Xiao, Xuefeng and Cui, Bin},
  booktitle={Proceedings of the IEEE/CVF International Conference on Computer Vision},
  pages={15982--15993},
  year={2025}
}

@inproceedings{zhaovidit,
  title={ViDiT-Q: Efficient and Accurate Quantization of Diffusion Transformers for Image and Video Generation},
  author={Zhao, Tianchen and Fang, Tongcheng and Huang, Haofeng and Wan, Rui and Soedarmadji, Widyadewi and Liu, Enshu and Li, Shiyao and Lin, Zinan and Dai, Guohao and Yan, Shengen and others},
  booktitle={The Thirteenth International Conference on Learning Representations},
  year={2025}
}

@article{feng2025q,
  title={Q-vdit: Towards accurate quantization and distillation of video-generation diffusion transformers},
  author={Feng, Weilun and Yang, Chuanguang and Qin, Haotong and Li, Xiangqi and Wang, Yu and An, Zhulin and Huang, Libo and Diao, Boyu and Zhao, Zixiang and Xu, Yongjun and others},
  journal={arXiv preprint arXiv:2505.22167},
  year={2025}
}

@article{wan2025wan,
  title={Wan: Open and advanced large-scale video generative models},
  author={Wan, Team and Wang, Ang and Ai, Baole and Wen, Bin and Mao, Chaojie and Xie, Chen-Wei and Chen, Di and Yu, Feiwu and Zhao, Haiming and Yang, Jianxiao and others},
  journal={arXiv preprint arXiv:2503.20314},
  year={2025}
}

@article{chen2025hunyuanvideo,
  title={HunyuanVideo-Avatar: High-Fidelity Audio-Driven Human Animation for Multiple Characters},
  author={Chen, Yi and Liang, Sen and Zhou, Zixiang and Huang, Ziyao and Ma, Yifeng and Tang, Junshu and Lin, Qin and Zhou, Yuan and Lu, Qinglin},
  journal={arXiv preprint arXiv:2505.20156},
  year={2025}
}

@article{li2025infinityhuman,
  title={InfinityHuman: Towards Long-Term Audio-Driven Human},
  author={Li, Xiaodi and Xie, Pan and Ren, Yi and Gan, Qijun and Zhang, Chen and Kong, Fangyuan and Yin, Xiang and Peng, Bingyue and Yuan, Zehuan},
  journal={arXiv preprint arXiv:2508.20210},
  year={2025}
}

@inproceedings{wang2025fantasytalking,
  title={Fantasytalking: Realistic talking portrait generation via coherent motion synthesis},
  author={Wang, Mengchao and Wang, Qiang and Jiang, Fan and Fan, Yaqi and Zhang, Yunpeng and Qi, Yonggang and Zhao, Kun and Xu, Mu},
  booktitle={Proceedings of the 33rd ACM International Conference on Multimedia},
  pages={9891--9900},
  year={2025}
}

@inproceedings{liu2025timestep,
  title={Timestep Embedding Tells: It's Time to Cache for Video Diffusion Model},
  author={Liu, Feng and Zhang, Shiwei and Wang, Xiaofeng and Wei, Yujie and Qiu, Haonan and Zhao, Yuzhong and Zhang, Yingya and Ye, Qixiang and Wan, Fang},
  booktitle={Proceedings of the Computer Vision and Pattern Recognition Conference},
  pages={7353--7363},
  year={2025}
}

@article{ma2025magcache,
  title={Magcache: Fast video generation with magnitude-aware cache},
  author={Ma, Zehong and Wei, Longhui and Wang, Feng and Zhang, Shiliang and Tian, Qi},
  journal={Advances in Neural Information Processing Systems},
  volume={38},
  pages={34348--34380},
  year={2026}
}

@inproceedings{liu2025speca,
  title={Speca: Accelerating diffusion transformers with speculative feature caching},
  author={Liu, Jiacheng and Zou, Chang and Lyu, Yuanhuiyi and Ren, Fei and Wang, Shaobo and Li, Kaixin and Zhang, Linfeng},
  booktitle={Proceedings of the 33rd ACM International Conference on Multimedia},
  pages={10024--10033},
  year={2025}
}

@inproceedings{zheng2025compute,
  title={Compute only 16 tokens in one timestep: Accelerating diffusion transformers with cluster-driven feature caching},
  author={Zheng, Zhixin and Wang, Xinyu and Zou, Chang and Wang, Shaobo and Zhang, Linfeng},
  booktitle={Proceedings of the 33rd ACM International Conference on Multimedia},
  pages={10181--10189},
  year={2025}
}

@article{guan2025forecasting,
  title={Forecasting When to Forecast: Accelerating Diffusion Models with Confidence-Gated Taylor},
  author={Guan, Xiaoliu and Jiang, Lielin and Chen, Hanqi and Zhang, Xu and Yan, Jiaxing and Wang, Guanzhong and Liu, Yi and Zhang, Zetao and Wu, Yu},
  journal={Knowledge-Based Systems},
  pages={114635},
  year={2025},
  publisher={Elsevier}
}

@inproceedings{liu2025reusing,
  title={From reusing to forecasting: Accelerating diffusion models with taylorseers},
  author={Liu, Jiacheng and Zou, Chang and Lyu, Yuanhuiyi and Chen, Junjie and Zhang, Linfeng},
  booktitle={Proceedings of the IEEE/CVF International Conference on Computer Vision},
  pages={15853--15863},
  year={2025}
}

@article{chen2025deltadit,
  title={\${\textbackslash}Delta\$-DiT: Accelerating Diffusion Transformers without Training via Denoising Property Alignment},
  author={Chen, Pengtao and Shen, Mingzhu and Ye, Peng and Cao, Jianjian and Tu, Chongjun and Bouganis, Christos-Savvas and Zhao, Yiren and Chen, Tao},
  journal={International Journal of Computer Vision},
  volume={134},
  number={6},
  pages={276},
  year={2026},
  publisher={Springer}
}

@article{heusel2017fid,
  title={Gans trained by a two time-scale update rule converge to a local nash equilibrium},
  author={Heusel, Martin and Ramsauer, Hubert and Unterthiner, Thomas and Nessler, Bernhard and Hochreiter, Sepp},
  journal={Advances in neural information processing systems},
  volume={30},
  year={2017}
}

@article{unterthiner2019fvd,
  title={FVD: A new metric for video generation},
  author={Unterthiner, Thomas and Van Steenkiste, Sjoerd and Kurach, Karol and Marinier, Rapha{\"e}l and Michalski, Marcin and Gelly, Sylvain},
  year={2019},
}

@article{lipmanflow,
  title={Flow matching for generative modeling},
  author={Lipman, Yaron and Chen, Ricky TQ and Ben-Hamu, Heli and Nickel, Maximilian and Le, Matt},
  journal={arXiv preprint arXiv:2210.02747},
  year={2022}
}

@article{kong2024hunyuanvideo,
  title={Hunyuanvideo: A systematic framework for large video generative models},
  author={Kong, Weijie and Tian, Qi and Zhang, Zijian and Min, Rox and Dai, Zuozhuo and Zhou, Jin and Xiong, Jiangfeng and Li, Xin and Wu, Bo and Zhang, Jianwei and others},
  journal={arXiv preprint arXiv:2412.03603},
  year={2024}
}

@article{lu2022dpm,
  title={Dpm-solver: A fast ode solver for diffusion probabilistic model sampling in around 10 steps},
  author={Lu, Cheng and Zhou, Yuhao and Bao, Fan and Chen, Jianfei and Li, Chongxuan and Zhu, Jun},
  journal={Advances in neural information processing systems},
  volume={35},
  pages={5775--5787},
  year={2022}
}

@inproceedings{zhang2018lpips,
  title={The unreasonable effectiveness of deep features as a perceptual metric},
  author={Zhang, Richard and Isola, Phillip and Efros, Alexei A and Shechtman, Eli and Wang, Oliver},
  booktitle={Proceedings of the IEEE conference on computer vision and pattern recognition},
  pages={586--595},
  year={2018}
}

@article{wang2004ssim,
  title={Image quality assessment: from error visibility to structural similarity},
  author={Wang, Zhou and Bovik, Alan C and Sheikh, Hamid R and Simoncelli, Eero P},
  journal={IEEE transactions on image processing},
  volume={13},
  number={4},
  pages={600--612},
  year={2004},
  publisher={IEEE}
}

@inproceedings{chung2016syncnet,
  title={Out of time: automated lip sync in the wild},
  author={Chung, Joon Son and Zisserman, Andrew},
  booktitle={Asian conference on computer vision},
  pages={251--263},
  year={2016},
  organization={Springer}
}

@inproceedings{hore2010psnr,
  title={Image quality metrics: PSNR vs. SSIM},
  author={Hore, Alain and Ziou, Djemel},
  booktitle={2010 20th international conference on pattern recognition},
  pages={2366--2369},
  year={2010},
  organization={IEEE}
}

@article{dao2022flashattention,
  title={Flashattention: Fast and memory-efficient exact attention with io-awareness},
  author={Dao, Tri and Fu, Dan and Ermon, Stefano and Rudra, Atri and R{\'e}, Christopher},
  journal={Advances in neural information processing systems},
  volume={35},
  pages={16344--16359},
  year={2022}
}

@article{wei2024aniportrait,
  title={Aniportrait: Audio-driven synthesis of photorealistic portrait animation},
  author={Wei, Huawei and Yang, Zejun and Wang, Zhisheng},
  journal={arXiv preprint arXiv:2403.17694},
  year={2024}
}

@inproceedings{zhang2025spargeattention,
  title={SpargeAttention: Accurate and Training-free Sparse Attention Accelerating Any Model Inference},
  author={Zhang, Jintao and Xiang, Chendong and Huang, Haofeng and Xi, Haocheng and Zhu, Jun and Chen, Jianfei and others},
  booktitle={Forty-second International Conference on Machine Learning},
  year={2025}
}

@article{Wang2023VideoLCMVL,
  title={Videolcm: Video latent consistency model},
  author={Wang, Xiang and Zhang, Shiwei and Zhang, Han and Liu, Yu and Zhang, Yingya and Gao, Changxin and Sang, Nong},
  journal={arXiv preprint arXiv:2312.09109},
  year={2023}
}

@inproceedings{meng2023distillation,
  title={On distillation of guided diffusion models},
  author={Meng, Chenlin and Rombach, Robin and Gao, Ruiqi and Kingma, Diederik and Ermon, Stefano and Ho, Jonathan and Salimans, Tim},
  booktitle={Proceedings of the IEEE/CVF conference on computer vision and pattern recognition},
  pages={14297--14306},
  year={2023}
}

@article{selvaraju2024forafastforwardcachingdiffusion,
  title={Fora: Fast-forward caching in diffusion transformer acceleration},
  author={Selvaraju, Pratheba and Ding, Tianyu and Chen, Tianyi and Zharkov, Ilya and Liang, Luming},
  journal={arXiv preprint arXiv:2407.01425},
  year={2024}
}

@inproceedings{konglet,
  title={Let Them Talk: Audio-Driven Multi-Person Conversational Video Generation},
  author={Kong, Zhe and Gao, Feng and Zhang, Yong and Kang, Zhuoliang and Wei, Xiaoming and Cai, Xunliang and Chen, Guanying and Luo, Wenhan},
  booktitle={The Thirty-ninth Annual Conference on Neural Information Processing Systems},
  year={2025}
}

@article{liu2025freqca,
  title={Freqca: Accelerating diffusion models via frequency-aware caching},
  author={Liu, Jiacheng and Cai, Peiliang and Zhou, Qinming and Lin, Yuqi and Kong, Deyang and Huang, Benhao and Pan, Yupei and Xu, Haowen and Zou, Chang and Tang, Junshu and others},
  journal={arXiv preprint arXiv:2510.08669},
  year={2025}
}

@inproceedings{budicache,
  title={DiCache: Let Diffusion Model Determine Its Own Cache},
  author={Bu, Jiazi and Ling, Pengyang and Zhou, Yujie and Wang, Yibin and Zang, Yuhang and Lin, Dahua and Wang, Jiaqi},
  booktitle={The Fourteenth International Conference on Learning Representations},
  year={2026}
}

@article{feng2025hicache,
  title={Hicache: Training-free acceleration of diffusion models via hermite polynomial-based feature caching},
  author={Feng, Liang and Zheng, Shikang and Liu, Jiacheng and Lin, Yuqi and Zhou, Qinming and Cai, Peiliang and Wang, Xinyu and Chen, Junjie and Zou, Chang and Ma, Yue and others},
  journal={arXiv preprint arXiv:2508.16984},
  year={2025}
}

@inproceedings{zhang2025blockdance,
  title={Blockdance: Reuse structurally similar spatio-temporal features to accelerate diffusion transformers},
  author={Zhang, Hui and Gao, Tingwei and Shao, Jie and Wu, Zuxuan},
  booktitle={Proceedings of the Computer Vision and Pattern Recognition Conference},
  pages={12891--12900},
  year={2025}
}

@inproceedings{chu2025omnicache,
  title={OmniCache: A Trajectory-Oriented Global Perspective on Training-Free Cache Reuse for Diffusion Transformer Models},
  author={Chu, Huanpeng and Wu, Wei and Feng, Guanyu and Zhang, Yutao},
  booktitle={Proceedings of the IEEE/CVF International Conference on Computer Vision},
  pages={16302--16312},
  year={2025}
}
\end{document}